\documentclass[final,5p,times,twocolumn]{elsarticle}

\usepackage{amssymb}
\usepackage{amsfonts}
\usepackage{graphicx}
\usepackage{stfloats}
\usepackage{caption}
\usepackage{color}
\usepackage{float}
\usepackage{threeparttable}
\usepackage{multirow}
\usepackage{booktabs}
\usepackage{adjustbox}
\usepackage{flushend,cuted}
\usepackage{subfig}
\usepackage{wrapfig} 
\biboptions{numbers,sort&compress} 

\usepackage{tikz}
\usepackage{comment}
\usepackage{amsmath,amssymb} 
\usepackage{color}

\usepackage{microtype}
\usepackage{bm}
\usepackage{mathtools} 

\usepackage{latexsym}
\usepackage{mathtools} 
\newcommand\scalemath[2]{\scalebox{#1}{\mbox{\ensuremath{\displaystyle #2}}}}  

\usepackage[ruled,vlined]{algorithm2e}
\usepackage{float} 
\usepackage{subfig}  

\let\today\relax
\makeatletter
\def\ps@pprintTitle{%
    \let\@oddhead\@empty
    \let\@evenhead\@empty
    \def\@oddfoot{\footnotesize\itshape
         {} \hfill\today}%
    \let\@evenfoot\@oddfoot
    }
\makeatother

\begin{document}
\date{}
\begin{frontmatter}

\title{Perception Improvement for Free: Exploring Imperceptible Black-box Adversarial Attacks on Image Classification}

\author[auth1]{Yongwei Wang}
\ead{yongweiw@ece.ubc.ca}
\author[]{Mingquan Feng}
\author[]{Rabab Ward}
\author[]{Z. Jane Wang}
\author[]{ Lanjun Wang}
\address[auth1]{Department
	of Electrical and Computer Engineering, University of British Columbia, Vancouver,
	BC, Canada}

\begin{abstract}
Deep neural networks are vulnerable to adversarial attacks. White-box adversarial attacks can fool neural networks with small adversarial perturbations, especially for large size images. However, keeping successful adversarial perturbations imperceptible is especially challenging for transfer-based black-box adversarial attacks. Often such adversarial examples can be easily spotted due to their unpleasantly poor visual qualities, which compromises the threat of adversarial attacks in practice. In this study, to improve the image quality of black-box adversarial examples perceptually, we propose structure-aware adversarial attacks by generating adversarial images based on psychological perceptual models. Specifically, we allow higher perturbations on perceptually insignificant regions, while assigning lower or no perturbation on visually sensitive regions. In addition to the proposed spatial-constrained adversarial perturbations, we also propose a novel structure-aware frequency adversarial attack method in the discrete cosine transform (DCT) domain. Since the proposed attacks are independent of the gradient estimation, they can be directly incorporated with existing gradient-based attacks. Experimental results show that, with the comparable attack success rate (ASR), the proposed methods can produce adversarial examples with considerably improved visual quality for free. With the comparable perceptual quality, the proposed approaches achieve higher attack success rates: particularly for the frequency structure-aware attacks, the average ASR improves more than 10\% over the baseline attacks. 
\end{abstract}


\end{frontmatter}

\section{Introduction}
\label{intro}

Deep neural networks (DNNs) have achieved significant progress in a wide range of machine learning tasks \cite{alexnet, resnet, ren2015faster, mask,revhashnet, fast_track}. However, their robustness has been greatly challenged by the existence of adversarial examples, where carefully perturbed images (as the inputs) can easily fool deep neural networks. Since Szegedy et al. \cite{Szegedy14} first reported adversarial examples, there have been intensive studies on the effectiveness of adversarial examples \cite{cw,mim,phyAttack,fgsm,pgd,jsma,asr19,dim}.  

In practice, a valid adversarial example satisfies two constraints: a) \textit{high attack success rate}, i.e., adversarial examples can fool the target models with a high attack success rate; and b) \textit{high perceptual quality}, i.e., adversarial examples are semantically preserving and meaningful, which indicates the image content is preserved and the image perceptual quality is as naturally-looking as possible.

\begin{figure}[ht]
\centering
\includegraphics[width=0.45\textwidth]{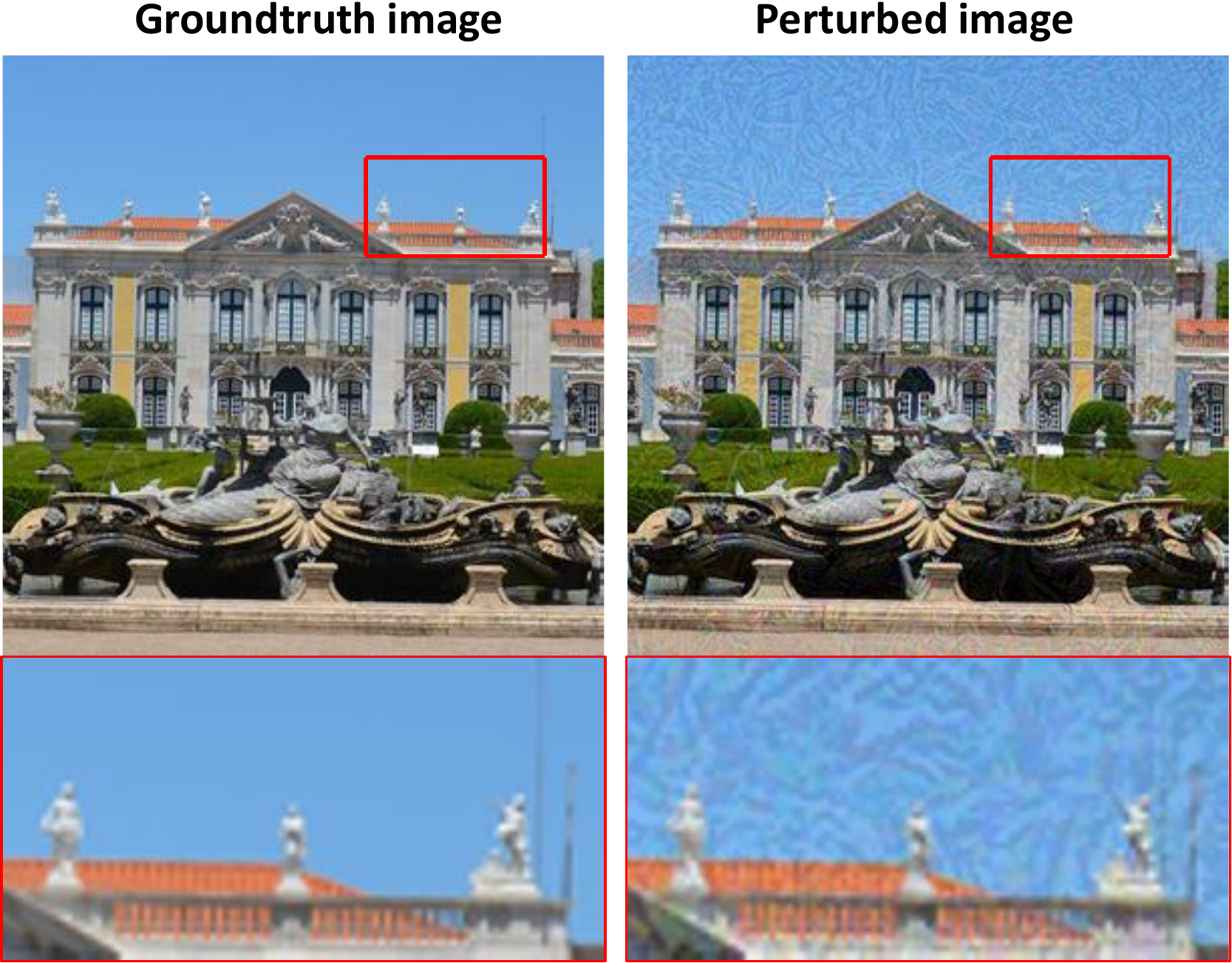} 
\caption{Illustration of a visually degraded adversarial example. Left: the groundtruth image; right: the perturbed image with adversarial perturbations generated by \cite{dim}. By zooming into the image patch (i.e, the red box), we can clearly notice the visual artifacts introduced by adversarial perturbations.}  
\label{fig:perturb_example}
\end{figure}

White-box adversarial attack methods~\cite{phyAttack,fgsm,pgd} can easily generate valid adversarial examples satisfying the above two constraints, because the adversary has full knowledge of the deployed model. However, to meet these constraints is much more challenging for black-box attacks \cite{mim,dim}. For example, \cite{dim}, one of the latest attacks with high attack success rates, requires relatively large perturbations, which can generally degrade the perceptual quality of the generated adversarial examples. For example in Fig.~\ref{fig:perturb_example}, we depict an adversarial example with perturbation generated by \cite{dim}, which displays unpleasant or unnatural visual artifacts. Despite those adversarial examples with poor visual qualities remain fooling the model, their threats to certain practical deployed systems (e.g., deepfake forensics \cite{li2020celeb}) can be largely compromised, because indeed they break the `imperceptibility' property of adversarial attacks and can be easily spotted and filtered out by sanity checks. As a result, a key problem we need to solve for black-box adversarial attacks is \textit{whether it is possible to keep a high attack success rate while preserving a naturally-looking visual quality?}     

According to studies on the human cognition system, we realize that the essence of visual degradation issues is the identical and independent perturbation bound for each pixel. More specifically, such identical and independent perturbation bound is incompatible with the human visual system, which is highly sensitive to the structural information in scene perception \cite{jnd95,jnd05}. In detail, structural representations can be described by edge, texture and luminance contrast by extracting oriented gradients and relative intensity from neighboring pixels in the spatial domain \cite{hvs93}. Moreover, visual frequency sensitivity can be integrated into constructing visual descriptors in the frequency domain \cite{zhang05,lin11}. Therefore, uniform distortions in previous adversarial attack studies are not aligned well with the human visual system. The visual quality issue is not obvious for white-box attacks because the perturbation bound can be very small due to the fully known information. However, we have to solve the visual quality issue in black-box attacks. 

To generate black-box adversarial examples by considering the human perception behavior is very challenging.  Firstly, to replace the uniform perturbation, we need a new type of distortion metric to represent the structural properties of images. In this study, we incorporate the results from psychological studies. The structure-aware image-dependent perceptual models \cite{jnd95} can identify which regions the visual systems pay more attention and which regions is more likely to be ignored.  These models have been applied in the fields of image compression \cite{hvs93} and video coding \cite{jnd05}, where higher compression rates are applied on the unnoticeable regions, but lower compression rates or even not to compress on the noticeable regions. We propose to leverage these perceptual models on setting a structure-aware adversarial attack. More specifically, we allow higher perturbations on perceptually insignificant regions, while assigning lower or no perturbation on significant regions.   

Secondly, only considering perturbations in the spatial domain is not enough, because perceptual systems are closely related to frequency selectivity \cite{lin11}. Although there exist frequency perceptual models to quantitatively measure frequency sensitivity (e.g., \cite{hvs93,zhang05}), it is nontrivial to incorporate frequency visual models to the adversarial attack setting. To leverage the frequency perceptual models, we propose to directly add adversarial perturbations in the frequency domain, i.e., we formulate a novel adversarial attack objective function in the frequency domain with the frequency sensitivity constraint, then frequency perturbation is conducted with gradients derived for each frequency sub-band.                    

Thirdly, there is always a trade-off between the attack success rate and perceptual quality \cite{eval19}.  Simply achieving imperceptible perturbations alone is not enough, while we still need to keep a high success rate in the black-box attacks.  In this study, we carefully select the structure-aware perceptual incorporation strategy to make them independent of the existing gradient-based attack algorithms. As a result, we can leverage the state-of-art gradient estimation methods, while constrain the perturbation setting based on the perceptual models.  

We summarize our major contributions as follows:
\begin{itemize}
    \item We design a framework to generate structure-aware distortions in adversarial attacks, and apply it on black-box adversarial attacks to preserve a naturally-looking visual quality while keeping a high attack success rate. Since the structure-aware strategy is independent of the gradient estimation, this framework can be generally extended to any gradient-based adversarial attack regardless of the white-box or black-box setting. 
    \item Besides the spatial structure-aware perturbations, we propose to incorporate the frequency perceptual models in the adversarial perturbation generation and we develop a novel structure-aware attack approach by adding adversarial perturbations in the frequency domain.
    \item Experiments demonstrate that, with the comparable attack success rate, the proposed methods have significant perceptual improvements when compared with the baseline attacks. Meanwhile, with the comparable perceptual quality, we also observe the improved attack success rate over the baseline attacks . 

\end{itemize}
\section{Background}
The existence of adversarial examples poses severe threats to deep learning models. A wide range of studies have been investigated to generate adversarial examples to fool neural networks with a high probability \cite{cw,mim,phyAttack,fgsm,pgd,jsma,asr19,dim,bim17}. However, all of these studies neglected the perceptual quality evaluation on adversarial examples. 

Meanwhile, limited attention has been paid to generating adversarial examples with high perceptual quality. Luo et al. introduced an overall noise sensitivity measure based on noise variance estimation for white-box attacks \cite{luo18}. Croce et al. introduced a sparse $\ell_0$ ball constraint to the query-limited attack and optimized the perturbation with local search \cite{croce19}. The sparse perturbations are assigned to sparse regions with high variances to reduce visual distortions. However, neither of them is aligned with human visual systems, because some complicated structures (e.g., textures, edges, luminance contrast) or frequency response of an image are not explicitly modeled.  

Furthermore, the idea of incorporating psychological studies \cite{jnd95,jnd05}  is inspired by related works from image compression \cite{hvs93} and video coding \cite{jnd05}. For instance, in video coding \cite{jnd05}, the perceptual-model incorporated codecs achieve both high perceptual fidelity and a high compression rate.

\subsection{Adversarial Attack Models}

Given a clean image $\boldsymbol{x}$, an image classifier $f_{\theta}$  predicts its label as $y$, i.e., $f_{\theta}(\boldsymbol{x}) = y$. Conventionally, a non-targeted adversarial example $\boldsymbol{x}^{*}$ can be formally defined as
\begin{equation}
\label{eq:regu}
    f_{\theta}(\boldsymbol{x^{*}}) \neq y, \quad \textrm{s.t.} \; \left \| \bm{x}^{*} - \bm{x} \right \|_p \le \epsilon
\end{equation}
By definition, an adversarial example $\boldsymbol{x}^{*}$ is bounded within the $\epsilon$ ball of $\boldsymbol{x}$, with distance measured by the $\ell_p$ norm.

With a higher $\epsilon$ factor, the attack methods produce relatively high attack success rate at the expense of possibly severely degraded perceptual quality. The key issue lies in the fact that the distortion criterion treats each pixel independently and assigns a uniform bound with each pixel. However, human eyes mainly perceive images using local and regional statistics. Therefore, the tolerable distortion level should be different from pixel to pixel due to their different structure information defined by neighboring regions \cite{hvs93,jnd05,lin11}.

\section{Perceptual Models}
The understanding of the human visual system is essential to generate high quality adversarial examples. We employ perceptual models to guide adversarial example generation. Perceptual models are developed over the years based on the property of human visual system over scene perception. Psychovisual study reveals that visual sensitivity relies on structural information rather than value changes at a single pixel \cite{hvs93,jnd95}. A common paradigm of perceptual models in image processing is the just-noticeable difference (JND) model, which was originally derived for image compression \cite{hvs93}. 

In JND models, the structure and local statistics are generally described by luminance sensitivity, contrast masking and frequency masking effects \cite{lin11}. There are various types of JND models in the literature, e.g., the spatial domain JND \cite{jnd99} and the frequency domain JND \cite{jnd95,jnd00,zhang05}. Based on JND models, we can estimate the maximal perturbation bounds within the imperceptibility constraint. It also indicates the perceptual importance on each pixel, and so we leverage it to design non-uniform distortions. We are motivated to incorporate the perceptual-model based constraint to generate adversarial examples with high visual quality. We briefly describe two JND models we adopt in the following sections.   

\subsection{Spatial JND Model}
In the spatial domain, we apply a basic JND model, which considers the image structure that consists of textures and local luminance distribution \cite{jnd05}. The JND profile is obtained by calculating the dominant values of its two structural components. Specifically, the spatial JND for a grayscale image, denoted by $JND_s$, is defined as follows: 
\begin{equation}
    \mathcal{JND}_s = \mathcal{TM} + \mathcal{LA}  - C \cdot \min \{ \mathcal{TM}, \mathcal{LA}\}
\end{equation}
where $\mathcal{TM}$ represents the texture masking, $\mathcal{LA}$ represents the luminance adaptation, and $C \in (0,1)$ measures the overlapping effect between the texture masking and luminance adaptation effects. Empirically, we set $C$ as 0.3 according to \cite{jnd05}. 

Texture masking refers to the ability of hiding or obscuring a superimposed stimulus with textures. \cite{jnd95,jnd05} show that the visual sensitivity to distortion is low in the texture-rich regions. The visual importance defined by texture masking is estimated as: 
\begin{equation}
    \mathcal{TM} = \max \limits_{k=1,2,3,4}  | \boldsymbol{x} * \boldsymbol{h}_k  | \cdot (\boldsymbol{m_{\boldsymbol{x}}} * \boldsymbol{l}_g)
\end{equation}
where $\boldsymbol{h}_k\; (k=1,2,3,4)$ are four directional high-pass filters for texture detection, $\boldsymbol{m}_{\boldsymbol{x}}$ denotes the edge map of image $\boldsymbol{x}$ given by the Canny edge detector \cite{canny}, and $\boldsymbol{l}_g$ represents a Gaussian low-pass filter. The filter parameters are provided in \ref{app:spatial_JND}. 

Compared with the absolute luminance of a single pixel, human perceptions are more sensitive to the relative luminance among its neighboring pixels. The luminance adaptation threshold is calculated based on Weber's law and deduced from psychological experiments under uniform background \cite{netravali2013digital}. The luminance adaptation effect $\mathcal{LA}$ is modeled as,      
\begin{equation}
   \mathcal{LA}_{i,j} =
  \begin{cases*}
    17 \times (  1 - \sqrt{\frac{\boldsymbol{\tilde{x}}_{i,j}}{127} } ) \quad& if $ \boldsymbol{\tilde{x}}_{i,j} \le 127 $ \\
    \frac{3 \times \left( \boldsymbol{\tilde{x}}_{i,j} - 127 \right)}{128}  + 3  & otherwise \\
  \end{cases*} 
\end{equation}
where $(i,j)$ denotes pixel position of a grayscale image, $\mathcal{LA}_{i,j}$ denotes the $(i,j)-$th component of the luminance adaptation map, $\boldsymbol{\tilde{x}} = \boldsymbol{x} * \boldsymbol{l}$, and $\boldsymbol{l}$ is a low-pass filter.

\subsection{Frequency JND Model}
In addition to spatial luminance adaptation and texture masking effects, the sensitivity of the human visual system is closely related to frequency sensitivity \cite{hvs93}. We adopt a frequency perceptual model proposed in \cite{zhang05}. To describe the frequency perceptual model in short, images are firstly decomposed into sub-band domains. Then, local contrast masking and spatial contrast sensitivity factors can be modeled based on frequency coefficients in each block. The final frequency JND is obtained as the multiplication of these two factors.       
\section{Methodology}

\subsection{Imperceptible Spatial-Domain Attack}
By departing from the identical $\ell_p$ bound for each pixel value, we consider the perceptual importance of pixels which directly depend on image local structures \cite{fgsm,dim}. Specifically, we allow larger perturbations to perceptually insignificant regions while smaller or no perturbations to perceptually significant regions. The perceptual importance is estimated from the neighborhood structures using the spatial JND model. To explicitly consider the imperceptibility property, we propose to incorporate and rectify existing adversarial example generation methods utilizing the spatial JND constraint.  

In the spatial domain, the optimization function of our JND-constraint spatial adversarial attack model is formulated as,
\begin{equation}
\label{eq:objs}
\begin{split}
    & f_{\theta}(\boldsymbol{x^{*}}) \neq y \\
    & \text{s.t.} \quad  |\boldsymbol{x} - \boldsymbol{x}^{*}| \preceq  \mathcal{JND}_s \\
\end{split}
\end{equation}
where $|\cdot|$ is the absolute value operator, $\mathcal{JND}_s$ denotes the spatial importance matrix estimated from the JND model computed from $\boldsymbol{x}$.  An intuitive explanation of our objective function is the following: We distinguish pixel-wise importance inherent in images extracted from local structures. Consequently the perturbation budgets vary from region to region. The gradient of the output with respect to the clean input $\boldsymbol{x}$ is a key value in the adversarial attack generation.  In black-box attacks, as there is no internal knowledge on either the model architecture or the loss function, it is impossible to calculate gradient directly. However, different types of black-box attacks leverage different methods to estimate such gradient information.  In the substitute model based attack which is our main focus, we can estimate the gradient with an substitute model \cite{fgsm,mim,dim}, then generate adversarial examples regarding to the new constraint as in Eq.(\ref{eq:objs}), and finally transfer examples to the black-box model.  In this study, we denote the gradient estimation method as $\boldsymbol{g}^{est}(\boldsymbol{x}, y)$.

The perceptual-constraint model can be solved using the gradient-based method iteratively,
\begin{equation}
\label{eq:sp_update}
    \boldsymbol{x}_{t+1}^{*} = \boldsymbol{x}_{t}^{*} +  \alpha \cdot \mathcal{JND}_s \odot \text{sign} (\boldsymbol{g}_{t}^{est}(\boldsymbol{x}_t^*, y))
\end{equation}
where $\odot$ denotes the elementwise product, $\boldsymbol{g}_{t}^{est}(\boldsymbol{x}_t^*, y)$ is the 
estimated gradient w.r.t. $\boldsymbol{x}_t$ at the $t-$th iteration. Informal studies show that exceeding JND thresholds occasionally does not yield severely degraded visibility. Therefore, we can exploit the additional tolerance by multiplying the importance map by a scalar factor $\alpha (\alpha \ge 1)$ in Eq.(\ref{eq:sp_update}). Then, we can better balance the trade-off between the attack success rate and image quality. 

Compared with the commonly used $\epsilon$ ball uniform bound, the proposed perturbation bound is image dependent and region dependent, which directly incorporates spatial perceptual models. In Eq.(\ref{eq:sp_update}), our image-dependent and stepsize-variant expression is a more general solution. Moreover, our method reduces to existing methods (e.g.,\cite{fgsm}) when we choose a uniform perturbation bound $\epsilon$ as $\alpha \cdot \max \{\mathcal{JND}_s\}$, then we have the same constraint optimization problem as in Eq.(\ref{eq:regu}) with an $\ell_{\infty}$ norm.     

The overall structure-aware adversarial spatial attack framework is illustrated in Algorithm 1. In this study, the JND threshold is calculated based on the grayscale version of a natural image. The final JND profile of a color image is formed by replicating the grayscale JND for each color channel. Although there exists color JND models, here we adopt a simple JND model in order to show the perceptual improvement by explicit utilization of structural information. 
\begin{algorithm}[h]
   \label{alg:alg1}
   \KwData{A black-box model $f(\boldsymbol{x})$, clean image $\boldsymbol{x}$, correct label $y$, gradient estimation method $\boldsymbol{g}^{est}(\boldsymbol{x}, y)$, scalar factor $\alpha_0$, and iteration number $T$. }
  \KwResult{An adversarial example $\boldsymbol{x^{*}}$.}
  Calculate $\mathcal{JND}_s$ from the grayscale version of $\boldsymbol{x}$. \\
  $\mathcal{JND} \leftarrow [\mathcal{JND}_s;\mathcal{JND}_s;\mathcal{JND}_s]$.\\
   Initialize $\boldsymbol{x}_{0}^{*} \leftarrow \boldsymbol{x}$, $t\leftarrow 0$ \\
   \While{$t<T$ and $f(\boldsymbol{x}^*_t)==y$}
    {estimate the gradient $\boldsymbol{g}_{t}^{est}(\boldsymbol{x}_t^*, y)$. \\
      $\boldsymbol{x}_{t+1}^{*} \leftarrow \boldsymbol{x}_{t}^{*} +  \frac{\alpha_0}{T} \cdot \mathcal{JND} \odot \text{sign} (\boldsymbol{g}_{t}^{est}(\boldsymbol{x}_t^*, y)) $, $t\leftarrow t+1$
    }
  $\boldsymbol{x}^{*} \leftarrow \boldsymbol{x}_{t}^{*}$. 
   \caption{The proposed spatial structure-aware (SSA) adversarial attack algorithm.} 
\end{algorithm}
\subsection{Imperceptible Frequency-Domain Attack}
Apart from the spatial domain perturbation, recently there were several pioneering works on perturbing images in the frequency domain. Tsuzuku el al. investigated the sensitivity of neural networks to certain Fourier basis functions based on the linearity hypothesis of neural networks \cite{freq19}. The adversarial examples can be crafted by making queries to the target model to find suitable Fourier basis. Adversarial examples from single Fourier attack method display repeated patterns in the pixel domain. Guo et al. restricted the adversarial perturbation space to the low frequency domain, and proposed a query-efficient attack method \cite{simBA}. Despite the effectiveness of low frequency perturbations, the visual quality of adversarial examples is significantly degraded \cite{sharma19}.    

In previous frequency attack methods, adversarial perturbations are added in the spatial domain with the uniform $\ell_p$-norm bound with frequency correction. However, our proposed frequency domain attack is directly conducted in the frequency domain iteratively without any spatial domain constraint. Instead we explicitly consider the perceptual distortion bounds in the frequency domain. This makes the proposed perceptual-constraint frequency attack different from existing adversarial attack methods. We observe that the proposed perceptual frequency-constraint adversarial attack generally yields higher perceptual quality than the spatial domain attack. 

In this study, we use the discrete cosine transform (DCT) domain for the frequency domain attack. For expression convenience, we consider the single channel case, since it is straightforward to extend operations to color images by performing transformations for each channel. Assume the clean image $\boldsymbol{x} \in \mathbb{R}^{N\times N}$, then we can obtain $\boldsymbol{X}$ by dividing the spatial image into square blocks of size $\mathcal{B} \times \mathcal{B}$. The DCT transform is conducted for each block $ \boldsymbol{x}^{b} (b=0,1,\cdots,\lceil \frac{N}{\mathcal{B}} \rceil -1 ) $ as, 
\begin{equation}
  \boldsymbol{X}^{b} = \boldsymbol{D}\boldsymbol{x}^{b} \boldsymbol{D}^{T}  
\end{equation}
where $\boldsymbol{D}$ is an orthogonal matrix, $\boldsymbol{D}\boldsymbol{D}^{T} = I_{\mathcal{B}\times\mathcal{B}}$, with entries $\boldsymbol{D}_{m,n} (m,n=0,1,\cdots,\mathcal{B}-1)$ as,
\begin{equation}
    \boldsymbol{D}_{m,n} = 
    \begin{cases*}
    \sqrt{\frac{1}{\mathcal{B}}} \quad & if $m = 0$ \\
    \sqrt{\frac{2}{\mathcal{B}}} cos(  \frac{(2m+1)n\pi}{2\mathcal{B}} ) \quad & otherwise \\
    \end{cases*}
\end{equation}
Similarly to the perceptual model-based spatial attack, we formulate the objective function for the frequency attack as,
\begin{equation}
\label{eq:objF}
\begin{split}
    & f_{\theta}(\boldsymbol{x^{*}}) \neq y \\
    & \text{s.t.} \quad  |\boldsymbol{X} - \boldsymbol{X}^{*}| \preceq  \mathcal{JND}_f \\
\end{split}
\end{equation}
where $\boldsymbol{X}, \boldsymbol{X}^{*}$ denote the clean and adversarial example in the frequency domain, respectively; $\mathcal{JND}_f$ refers to the JND matrix estimated by the frequency JND model.  

To solve Eq.(\ref{eq:objF}), we need to calculate the gradient of the loss function w.r.t. $\boldsymbol{X}$. At each block, the gradient w.r.t. each frequency coefficient $\boldsymbol{X}_{u,v}^{b} (u,v=0,1,\cdots,\mathcal{B}-1)$ can be calculated by propagating spatial gradient to the DCT domain,
\begin{equation}
    \boldsymbol{G}^{est}(\boldsymbol{X}^b_{u,v},y) = \sum \limits_{i=1}^{\mathcal{B}} \sum \limits_{j=1}^{\mathcal{B}} \boldsymbol{g}^{est}(\boldsymbol{x}^b_{i,j},y) \cdot \frac{\partial \boldsymbol{x}_{i,j}^{b}}{\partial \boldsymbol{X}_{u,v}^{b}}
\end{equation}
And we derive the gradient propagation in the matrix form,
\begin{equation}
\label{eq:Grad}
\boldsymbol{G}^{est}(\text{Vec }\boldsymbol{X}^b)  = \boldsymbol{g}^{est}(\text{Vec }\boldsymbol{x}^b) ^{T} \cdot \left( \boldsymbol{D}^T \otimes \boldsymbol{D}^T  \right)
\end{equation}
where $Vec$ denotes the matrix vectorization operation, $\cdot$ and $\otimes$ denote  inner product and matrix Kronecker product, respectively. Finally, we obtain the frequency gradient estimation $\boldsymbol{G}^{est}$.

With the frequency coefficient gradient computed from Eq.(\ref{eq:Grad}) as $\boldsymbol{G}_t^{est}$ at the $t-$th iteration, we can readily perform frequency attack with frequency JND in the DCT domain, 
\begin{equation}
    \label{eq:freq}
    \boldsymbol{X}_{t+1}^{*} = \boldsymbol{X}_{t}^{*} +  \beta \cdot \mathcal{JND}_f \odot \boldsymbol{G}_{t}^{est}
\end{equation}
where $\boldsymbol{X}_{t}^{*}$ denotes adversarial example in the DCT domain at iteration $t \; (t=1,2,\cdots,T)$, $\beta=\beta_0/T$, $\beta_0$ is a scalar factor of frequency JND to balance the compromise between perceptual quality and attack success rates.

Finally, the structure-aware frequency perturbation method is described in detail in Algorithm 2.  

\begin{algorithm}[tb]
   \caption{The proposed frequency structure-aware (FSA) adversarial attack algorithm.} 
   \label{alg:alg2}
   \KwData{ A black-box model $f(\boldsymbol{x})$, clean image $\boldsymbol{x}$, correct label $y$, gradient estimation method $\boldsymbol{g}^{est}(\boldsymbol{x}, y)$, scalar factor $\beta_0$, and iteration number $T$.}
    \KwResult{An adversarial example $\boldsymbol{x^{*}}$. }
    Calculate $\mathcal{JND}_f$ from DCT coefficients of grayscale version of $\boldsymbol{x}$. \\ 
    $\mathcal{JND} \leftarrow [\mathcal{JND}_f;\mathcal{JND}_f;\mathcal{JND}_f]$. \\
    Initialize: $\boldsymbol{x}_{0}^{*} \leftarrow \boldsymbol{x}$, $\boldsymbol{X}_{0}^{*} \leftarrow DCT(\boldsymbol{x})$, $t\leftarrow 0$. \\
 \While{$t<T$ and $f(\boldsymbol{x}^*_t)==y$}
   { Estimate the spatial gradient as $\boldsymbol{g}_{t}^{est}(f_{\theta}, \boldsymbol{x}_t^*, y)$. \\
    Calculate gradient w.r.t. DCT coefficient using Eq.(\ref{eq:Grad}) as $\boldsymbol{G}_t^{est}$ \\
     $\boldsymbol{X}_{t+1}^{*} \leftarrow \boldsymbol{X}_{t}^{*} +  \frac{\beta_0}{T} \cdot \mathcal{JND} \odot \boldsymbol{G}_{t}^{est} $\\
    $\boldsymbol{x}^*_{t+1} \leftarrow iDCT(\boldsymbol{X}^*_{t+1})$, $t \leftarrow t+1$
}
$\boldsymbol{x}^{*} \leftarrow \boldsymbol{x}^*_{t}$ 
\end{algorithm}

\section{Experimental results}
In this section, we evaluate the proposed structure-aware algorithms on three baseline attacks: Fast Sign Gradient Method(FGSM) \cite{fgsm},  Momentum Iterative Method(MIM) \cite{mim}, and Diverse Inputs Method(DIM) \cite{dim}. Firstly, we describe the experimental setup and introduce the quantitative visual metrics that we adopted in the comparison. We then experimentally demonstrate the superiority of the proposed methods over baselines on the perceptual quality and the attack success rate.  The perturbation residues are illustrated to show the structure-aware property. Finally, we discuss the sensitivity of the parameters in the methods.  

\subsection{Experimental Setup}
For substitute model based attacks, the substitute model is a cleanly trained Inc-v3 model \cite{incv3} provided by the PyTorch pretrained model zoo \cite{pytorch}.  We evaluate the effectiveness of the adversarial examples on six models, three of which are cleanly trained, i.e,. Inc-v4 \cite{incv4}, ResNet-101, ResNet-152 \cite{resnet}, and the rest three models are adversarially trained, i.e., $\textrm{Inc-v3}_{\textrm{adv}}$, $\textrm{Inc-v3}_{\textrm{ens3}}$, $\textrm{Inc-v3}_{\textrm{ens4}}$ \cite{advTrain}. These models are from the NeurIPS 2017 competition track on adversarial attacks \cite{escalera2018nips}. 

For the dataset, we randomly choose 1000 images from the ImageNet validation dataset \cite{imagenet}, which can be correctly classified by the six evaluation models in the substitute model attack setting. The uniform perturbation bound $\epsilon$ is set as 14 to have a good attack success rate. The maximum iteration number $T$ is set as 10, which is a default parameter in existing studies \cite{mim,dim}. For DCT transformation, we set block size as $8 \times 8$, as commonly used in JPEG compression and video coding \cite{zhang05}. 

\subsection{Evaluation Metrics}

To reliably evaluate the perceptual improvement of the proposed structure-aware attacks,  we adopt four image quality assessment (IQA) metrics: multiSim3 \cite{wang2003multiscale}, Feature Similarity for color images (FSIMc) \cite{zhang2011fsim}, Natural Image Quality Evaluator (NIQE) \cite{mittal2012making} and Mean Opinion Score (MOS) \cite{streijl2016mean}. MultiSim3 and FSIMc are full-reference IQA metrics, with scores within [0,1] where a higher score indicates better visual quality. NIQE is a no-reference IQA metric to measure the naturalness of tested images. NIQE produces a non-negative value, where lower values suggest better naturalness. MOS is a popular human subjective test where we adopt the absolute category rating principle, with image quality score ranging from 1 to 5. The higher the MOS, the better images appears visually similar to clean images. The detailed setting of our MOS test is in \ref{mos_setting}. 

To evaluate the attack effectiveness, we employ the averaged attack success rates (ASR) on six victim models \cite{escalera2018nips}. In the following sections, for simplicity, we term structure-aware approaches as SSA (Spatial-Structure-Aware) and FSA (Frequency-Structure-Aware). Since our proposed methods are independent of gradient estimation methods, we individually incorporate the structure-aware strategies to different gradient-based baseline attacks in the following sections. 

\subsection{Perception Improvement Assessment}
\label{sec:perceptual_improvement}
In this section, we compare the perceptual quality between three baseline attacks \cite{fgsm,mim,dim} and our proposed ones, respectively. In each comparison,  we firstly keep the average ASRs comparable between the baseline and the proposed ones, i.e., the proposed methods produce equal or slightly higher ASR than the baselines. Then, we provide both quantitative and qualitative comparison results on generated adversarial examples.

\textbf{Comparison with FGSM Attack \cite{fgsm}:} The Fast Sign Gradient Method (FGSM) is a one-step gradient-based attack method, which is a fundamental and widely adopted attack method. The perturbation is generated by maximizing the loss (e.g. cross-entropy) function $J(f_{\theta}, \boldsymbol{x}, y)$ w.r.t. the input image. The FGSM method meets $\left \| \boldsymbol{x} - \boldsymbol{x^{*}}  \right \| \le \epsilon$, and it has an expression as,          
\begin{equation}
    \boldsymbol{x}^{*} = \boldsymbol{x} + \epsilon \cdot \text{sign} (\boldsymbol{g})
\end{equation}
where $\boldsymbol{g} = \nabla_{\boldsymbol{x}} J(f_{\theta}, \boldsymbol{x}, y)$ denotes the gradient of the loss function w.r.t the clean sample. 

Table \ref{tab: fgsmASR1} shows the attack success rates of FGSM and our proposed variants, i.e., SSA-FGSM and FSA-FGSM. To have a comparable attack success rate, we choose $\alpha_0=2.2, \; \beta_0=50$. This table shows that SSA-FGSM has similar attack success rate with FGSM for both cleanly trained models and adversarially trained models, while FSA-FGSM approach gives superior attack success rate for adversarially trained models than cleanly trained ones. For a fair comparison, we keep their averaged ASR comparable as: 27.8$\%$ (FGSM), 27.8$\%$ (SSA-FGSM) and 29.8$\%$ (FSA-FGSM), respectively. 
\begin{table}[ht]
\caption{Attack success rate comparisons between FGSM and the proposed SSA-FGSM and FSA-FGSM methods. The attack success rate is in percent ($\%$).}
\centering
\begin{adjustbox}{width=0.499\textwidth}
\begin{tabular}{cccccccc}
\hline\noalign{\smallskip}
Attack   & ResNet-101 & ResNet-152 & Inc-v4 & $\textrm{Inc-v3}_{\textrm{adv}}$ & $\textrm{Inc-v3}_{\textrm{ens3}}$ & $\textrm{Inc-v3}_{\textrm{ens4}}$ & Avg ASR \\
\noalign{\smallskip}
\hline
\noalign{\smallskip}
FGSM     & 33.2      & 32.0      & \textbf{35.3}  & 22.7     & 25.3      & 18.1      & 27.8    \\ 
\textbf{SSA-FGSM} & \textbf{34.9}      & \textbf{34.4}      & 34.2  & 21.8     & 25.4      & 16.1      & 27.8    \\ 
\textbf{FSA-FGSM} & 28.4      & 27.2      & 29.7  & \textbf{25.5}     & \textbf{31.7}      & \textbf{36.4}      & \textbf{29.8}    \\ 
\hline
\end{tabular}
\end{adjustbox}
\label{tab: fgsmASR1}
\end{table}

Then we quantitatively assess the \textit{visual superiority} of the proposed methods in Table \ref{tab: fgsmVisual}. For IQA metrics, i.e., multiSim3, NIQE, FSIMc and MOS, the proposed SSA-FGSM achieves improvement by 3.4\%, 5.2\%, 0.45 and 1.09; and the proposed FSA-FGSM improves four IQA metrics by: 7.5\%, 15.3\%, 1.44, and 2.0, respectively. The quantitative comparison results confirm the significant perceptual improvement of the proposed methods over the vanilla FGSM attack. It is worthy to note that such visual improvement is obtained for free since we directly incorporate our strategies into vanilla FGSM. More importantly, compared with vanilla FGSM, the proposed methods require no sacrifice of the attack performance (i.e., average attack success rates). 

\begin{table}[ht]
\caption{Visual quality comparisons between FGSM, SSA-FGSM and FSA-FGSM methods. The symbol ``$\uparrow$'' (``$\downarrow$'') indicates that a higher (lower) value is better in perceptual quality.} 
\centering
\begin{adjustbox}{width=0.43\textwidth}
\begin{tabular}{ccccc}
\hline\noalign{\smallskip}
Attack   & multiSim3 ($\uparrow$) & FSIMc ($\uparrow$) & NIQE ($\downarrow$)  & MOS ($\uparrow$) \\
\noalign{\smallskip}
\hline
\noalign{\smallskip}
FGSM     & 0.862      & 0.762      & 3.002   &    1.79     \\ 
\textbf{SSA-FGSM} & 0.896      & 0.814      & 2.664  &   2.88    \\ 
\textbf{FSA-FGSM} & \textbf{0.937}      & \textbf{0.915 }   &  \textbf{1.560}  &  \textbf{3.79}  \\ 
\hline
\end{tabular}
\end{adjustbox}
\label{tab: fgsmVisual}
\end{table}

\textbf{Comparison with MIM Attack \cite{mim}:} To improve the adversarial transferability, momentum is introduced to obtain the iterative version of FGSM as Momentum Iterative Method (MIM):
\begin{equation}
    \boldsymbol{x}_{t+1}^{*} = \boldsymbol{x}_{t}^{*} + \frac{\epsilon}{T} \cdot \text{sign} (\boldsymbol{g}_{t+1})
\end{equation}
where $T$ denotes the number of iterations, and the accumulated gradient is updated as, $\boldsymbol{g}_{t+1} = \mu \cdot \boldsymbol{g}_{t} + \frac{\nabla_{\boldsymbol{x}} J(f_{\theta, \boldsymbol{x}_{t}^{*},y})}{\left \|  \nabla_{\boldsymbol{x}} J(f_{\theta, \boldsymbol{x}_{t}^{*},y}) \right \|_1}$. After getting the updated gradient, we incorporate structural-aware strategies into MIM, and obtain the proposed SSA-MIM and FSA-MIM methods. We use $\mu=1.0$ as suggested in \cite{mim}.    
\begin{table}[ht]
\caption{Attack success rate comparisons between MIM and the proposed SSA-MIM and FSA-MIM methods. The attack success rate is in percent ($\%$).} 
\centering
\begin{adjustbox}{width=0.499\textwidth}
\begin{tabular}{cccccccc}
\hline\noalign{\smallskip}
Attack   & ResNet-101 & ResNet-152 & Inc-v4 & $\textrm{Inc-v3}_{\textrm{adv}}$ & $\textrm{Inc-v3}_{\textrm{ens3}}$ & $\textrm{Inc-v3}_{\textrm{ens4}}$ & Avg ASR \\
\noalign{\smallskip}
\hline
\noalign{\smallskip}
MIM     & \textbf{46.8}      & 44.8      & 56.0  & 24.8     & 29.3      & 30.0      & 38.6    \\
\textbf{SSA-MIM} & 44.2      & \textbf{46.0}      & \textbf{56.1}  & 26.1     & 30.3      & 29.8      & 38.8    \\
\textbf{FSA-MIM} & 40.7      & 39.2      & 47.0  & \textbf{34.3}     & \textbf{34.5}      & \textbf{41.2}      & \textbf{39.5}    \\
\hline
\end{tabular}
\end{adjustbox}
\label{tab: mimASR}
\end{table}

In Table \ref{tab: mimASR}, we compare the attack success rates of the MIM method and the proposed variants, e.g., SSA-MIM and FSA-MIM methods. The parameters for the two methods are $\alpha_0=2.3, \; \beta_0=6.0$ for a comparable attack success rate with respect to the vanilla MIM method, i.e., the averaged attack success rates are 38.6$\%$ for MIM, 38.8$\%$ for SSA-MIM, and 39.5$\%$ for FSA-MIM, respectively.

The quantitative IQA results are computed and reported in Table~\ref{tab: mimVisual}. Overall, SSA-MIM improves four metrics individually and FSA-MIM achieves even more improved perceptual qualities. Specifically, the quantitative IQA improvements are: 4.3\% on multiSim3, 8.3\% on FSIMc, 0.83 on NIQE and 1.68 on MOS, respectively. Therefore, the perceptual qualities of vanilla MIM can be largely improved by the utilization of the proposed structural-aware approaches. 
\begin{table}[ht]
\caption{Visual quality comparisons between MIM, SSA-MIM and FSA-MIM methods. The symbol ``$\uparrow$'' (``$\downarrow$'') indicates that a higher (lower) value is better in perceptual quality.}
\centering
\begin{adjustbox}{width=0.43\textwidth}
\begin{tabular}{ccccc}
\hline\noalign{\smallskip}
Attack   & multiSim3 ($\uparrow$) & FSIMc ($\uparrow$) & NIQE ($\downarrow$)  & MOS ($\uparrow$) \\
\noalign{\smallskip}
\hline
\noalign{\smallskip}
MIM     & 0.905      & 0.815       & 2.398  &     2.12    \\ 
\textbf{SSA-MIM} & 0.927      & 0.855      & 2.058  &  3.17     \\ 
\textbf{FSA-MIM} & \textbf{0.948}      & \textbf{0.928 }      & \textbf{1.569}  &  \textbf{3.80}  \\ 
\hline
\end{tabular}
\end{adjustbox}
\label{tab: mimVisual}
\end{table}

\textbf{Comparison with DIM Attack \cite{dim}:} In the DIM method, the inputs to the model are stochastically transformed copies of the original image to increase the adversarial transferability. At each iteration, correspondingly the gradient is updated with the transformation with a probability $p$. In the experiments, we selected $p$ as 0.7 which was reported to achieve the highest averaged attack success rates \cite{dim}. Based on the DIM method, we have derivations of our proposed structure-aware variants, i.e., SSA-DIM and FSA-DIM methods.     

\begin{table}[ht]
\caption{Attack success rate comparisons between DIM and the proposed SSA-DIM and FSA-DIM methods. The attack success rate is in percent ($\%$).} 
\centering
\begin{adjustbox}{width=0.499\textwidth}
\begin{tabular}{cccccccc}
\hline\noalign{\smallskip}
Attack   & ResNet-101 & ResNet-152 & Inc-v4 & $\textrm{Inc-v3}_{\textrm{adv}}$ & $\textrm{Inc-v3}_{\textrm{ens3}}$ & $\textrm{Inc-v3}_{\textrm{ens4}}$ & Avg ASR \\
\noalign{\smallskip}
\hline
\noalign{\smallskip}
DIM     & \textbf{64.2}      & 62.8      & 73.6  & 31.6     & 32.6      & 32.1      & 49.5    \\
\textbf{SSA-DIM} & 62.8      & \textbf{63.1}      & \textbf{73.7}  & 33.6     & 35.3      & 33.4      & 50.3    \\
\textbf{FSA-DIM} & 53.0      & 52.2      & 60.5  & \textbf{45.3}     & \textbf{43.0}      & \textbf{49.2}      & \textbf{50.5}    \\
\hline
\end{tabular}
\end{adjustbox}
\label{tab: dimASR}
\end{table}

To make the proposed attacks comparable with DIM \cite{dim} in ASR, we adopt $\alpha_0=2.35, \; \beta_0=6.5$ and show the attack success rate comparison between DIM, SSA-DIM and FSA-DIM methods in Table \ref{tab: dimASR}. The averaged attack success rates are 49.5$\%$, 50.3$\%$ and 50.5$\%$, respectively.  

\begin{table}[ht]
\caption{Visual quality comparisons between DIM, SSA-DIM and FSA-DIM methods. The symbol ``$\uparrow$'' (``$\downarrow$'') indicates that a higher (lower) value is better in perceptual quality.}
\centering
\begin{adjustbox}{width=0.43\textwidth}
\begin{tabular}{ccccc}
\hline\noalign{\smallskip}
Attack   & multiSim3 ($\uparrow$) & FSIMc ($\uparrow$) & NIQE ($\downarrow$)  & MOS ($\uparrow$) \\
\noalign{\smallskip}
\hline
\noalign{\smallskip}
DIM     & 0.906      & 0.816      & 2.431   &       2.15  \\ 
\textbf{SSA-DIM} & 0.926      & 0.851     & 2.112   &    3.15   \\ 
\textbf{FSA-DIM} & \textbf{0.941}      & \textbf{0.921}      & \textbf{1.684}  &  \textbf{3.58}  \\ 
\hline
\end{tabular}
\end{adjustbox}
\label{tab: dimVisual}
\end{table}

Compared with vanilla MIM (Table \ref{tab: mimASR}), vanilla DIM improves the averaged ASR by about 10\% (Table \ref{tab: dimASR}). Correspondingly, we observe that the proposed attacks (i.e. SSA-DIM and FSA-DIM) also improve their ASRs over MIM-based methods (i.e. SSA-MIM and FSA-MIM) by a similar margin. This observation confirms that our proposed structure-aware strategies are indeed independent of gradient-based methods, i.e., the incorporation of perceptual models into existing attacks can still maintain their attack ability. 

Meanwhile, we notice that vanilla DIM also suffers from the visual quality problem as reported in Table \ref{tab: dimVisual}, e.g., the FSIMc is only 0.816. By contrast, SSA-DIM improves the metric by 3.9\% and FSA-DIM further boosts its FSIMc metric to be 0.921.  

\begin{figure*}[ht]
\centering
\includegraphics[width=18cm,height=14.4cm]{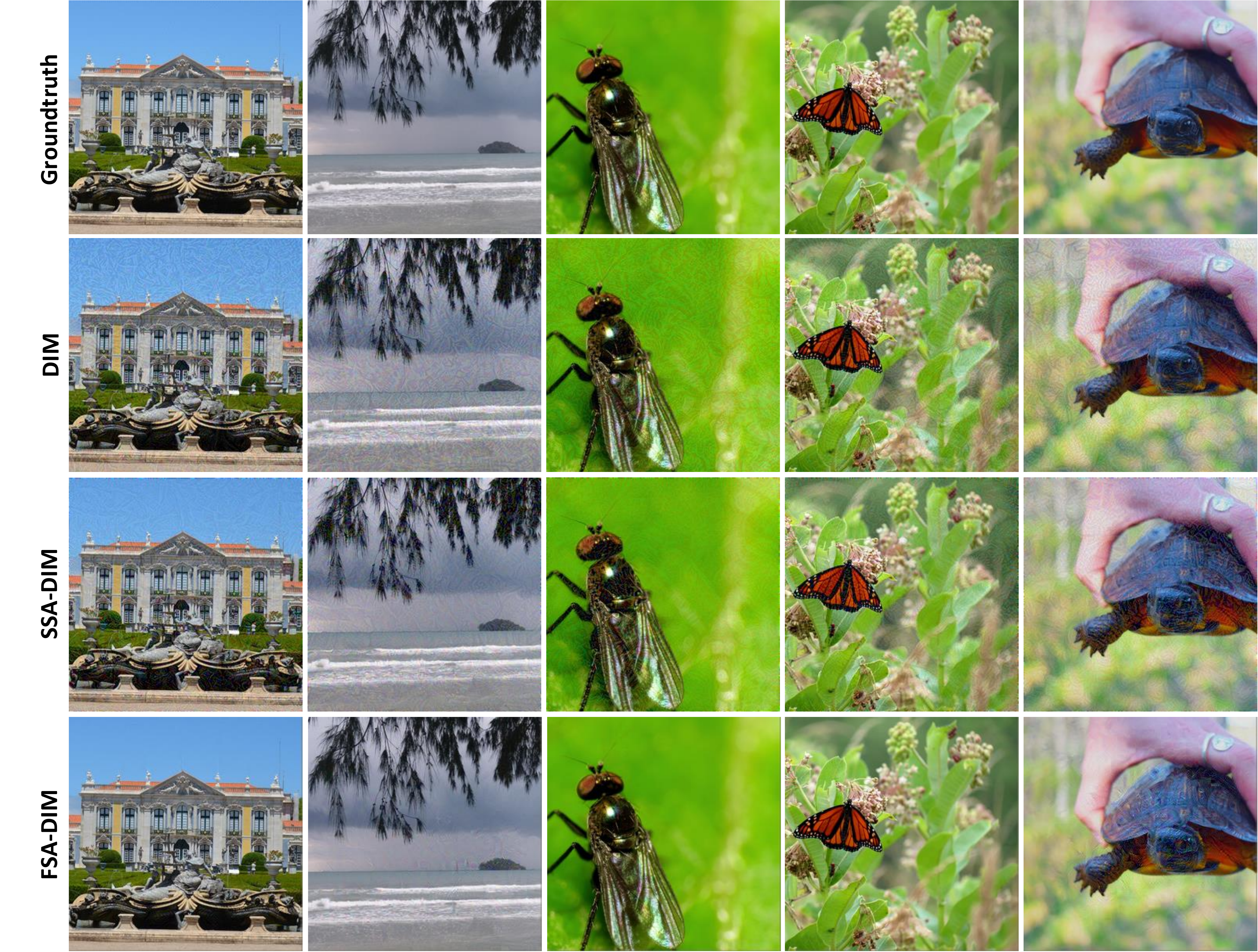}
\caption{Examples of perceptual image quality comparison between DIM, SSA-DIM and FSA-DIM methods. We recommend to zoom into the digital version for better visual comparison.}  
\label{fig:dim_visual}
\end{figure*}

Finally, we show several typical adversarial examples for qualitative visual comparison in Fig.~\ref{fig:dim_visual}. The first row depicts the clean images, and the last three rows display adversarial examples generated from DIM, SSA-DIM and FSA-DIM, respectively. For DIM, we observe the perceptual degradation phenomenon in adversarial images, especially in the smooth regions. In detail, the texture-like distortions make adversarial examples visually unpleasant and easily to be spotted (please zoom in Fig.~\ref{fig:dim_visual} for better comparison). Compared with DIM, SSA-DIM clearly improves the perceptual quality by re-allocating larger perturbation budgets to those visual insensitive regions based on spatial perceptual models. In the last row, we observe that the proposed FSA-DIM produces adversarial examples with almost imperceptible visual quality. Therefore, with the proposed structure-aware strategies (i.e., SSA and FSA), we can achieve comparable attack success rates yet with significantly higher visual quality over baseline methods, both quantitatively and qualitatively. 

In conclusion, this section shows experimental results and compare the perceptual improvement of the proposed structure-aware attacks with baseline attacks respectively. Overall, both the proposed spatial perceptual and frequency perceptual approaches can clearly improve the visual quality of adversarial examples with comparable average ASRs. Particularly for the frequency perceptual attacks, our proposed methods can generate almost imperceptible adversarial examples for each compared baseline attack. 

\subsection{ASR Improvement Assessment}
\label{sec:asr_assess}
In general, for the same attack, we can always maintain better visual quality (with less adversarial perturbations) at the expense of lower attack success rates \cite{mim}. In this section, we compare ASRs of three baseline methods and our proposed methods. To be specific, we decrease the perturbation budget $\epsilon$ of each baseline attack to make their IQA metrics comparable with the proposed methods individually. The IQA values of SSA and FSA have been reported in Table \ref{tab: fgsmVisual} - Table \ref{tab: dimVisual} as the comparison reference.  

\begin{table}[ht]
\caption{ASR improvement comparisons between the baseline attacks and their SSA/FSA versions, with the comparable visual quality.} 
\centering
\begin{adjustbox}{width=0.4\textwidth}

\begin{tabular}{|c|c|c|c|c|}
\hline
\multirow{2}{*}{Attacks} & \multicolumn{2}{c|}{SSA equivalent} & \multicolumn{2}{c|}{FSA equivalent} \\ \cline{2-5} 
                               & FSIMc      & $\Delta$ASR (\%)           
                               & FSIMc      & $\Delta$ASR (\%)   \\ \hline
FGSM                         & 0.801      & 1.9            & 0.913      & 10.1      \\ \hline
MIM                           & 0.851      & 3.0           & 0.924      & 11.5      \\ \hline
DIM                         & 0.851      & 4.5          & 0.923      & 13.1      \\ \hline
\end{tabular}

\end{adjustbox}
\label{tab: ASR_improve}
\end{table}

With comparable visual quality (e.g. FSIMc), we show the ASR improvement $\Delta$ASR in Table \ref{tab: ASR_improve}. For the spatial perception-based methods, ASR improvement ranges from 1.9\% to 4.5\%. For the frequency perception-incorporated methods, ASR improves over baselines by 10.1\% to 13.1\%. This comparison result reveals another superiority of the proposed methods: by incorporating the proposed structure-aware strategies, we can achieve higher ASRs than baselines with comparably good visual quality. Therefore we can conclude that, compared with baseline attacks, the proposed methods manage to obtain a better trade-off between the attack success rates and perceptual quality.         

\subsection{Perturbation Residues} 
\label{sec:residues}
\begin{figure}[ht] 
\centering
\includegraphics[width=0.55\textwidth]{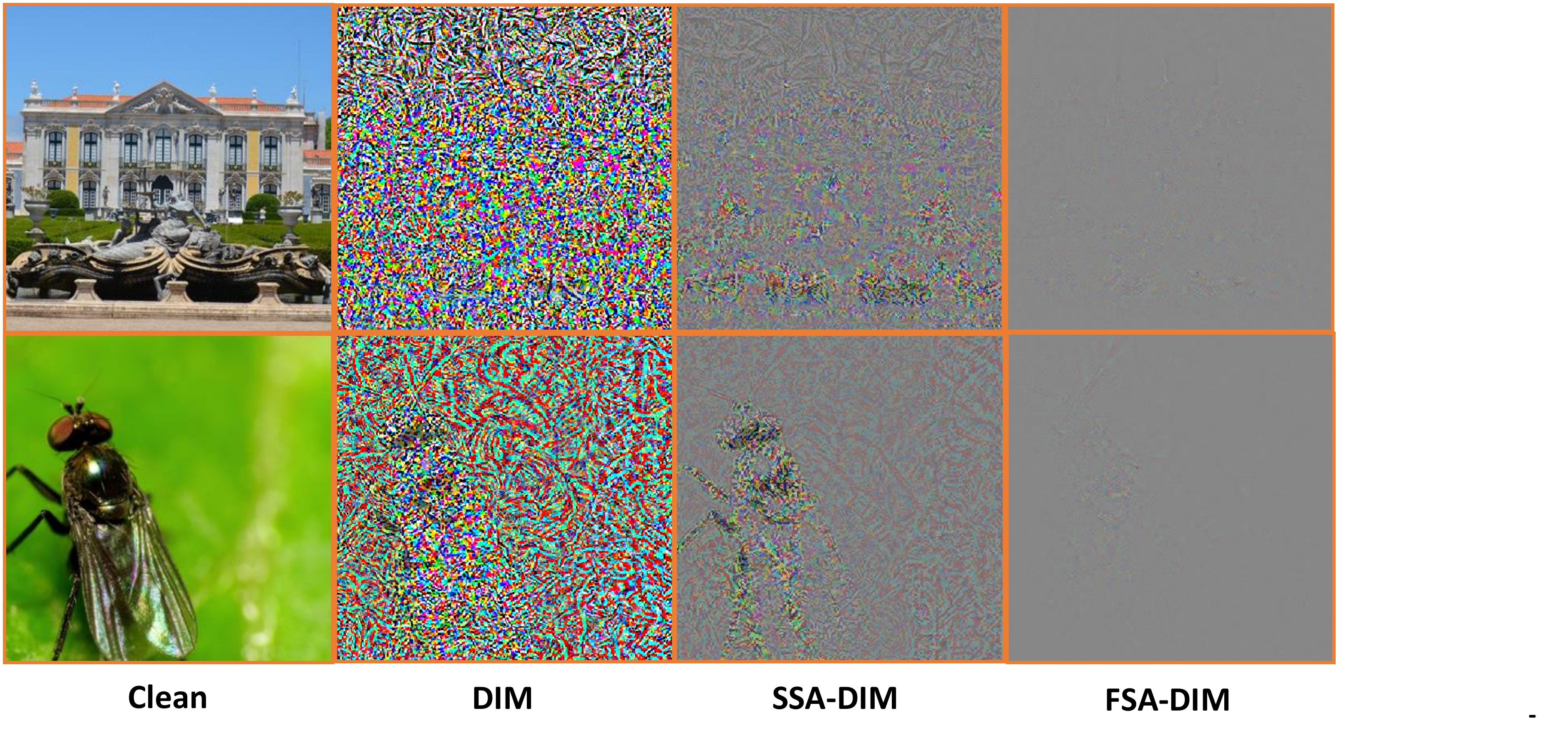}  
\caption{Comparison of perturbation residues between DIM ($\epsilon=14$), SSA-DIM ($\alpha_0=2.35$) and FSA-DIM ($\beta_0=6.5$) attacks on example images. }  
\label{fig:dim_residues}
\end{figure}

To better understand the perceptual-based attacks, as an example we visualize perturbation residues of the DIM-based methods in Fig.~\ref{fig:dim_residues}. The parameters of attacks are the same as in Table~\ref{tab: dimVisual}. In general, we observe that DIM uniformly perturbs all pixels of the image which accounts for the visual degradation issue. By contrast, SSA-DIM mainly perturbs the visual insignificant regions which can be computed from spatial perceptual structures. Meanwhile,  FSA-DIM approach perturbs the clean images with frequency insensitive adversarial perturbations in frequency perceptual bands, which generally appears invisible in the spatial domain.       

\subsection{Parameter Sensitivity} 
In this section, we study the effect of hyperparameters $\epsilon, \; \alpha_0$ and $\beta_0$ in the proposed attacks. To better illustrate the comparison trend of visual quality with respect to different hyperparameters, we normalize the NIQE values to be $\textrm{NIQE}'$: $NIQE'=1 - NIQE/NIQE_{ub}$, where $NIQE_{ub}$ is an upper bound of $NIQE$ values for all experiments we conducted. A higher $\textrm{NIQE}'$ value indicates the better visual quality or vice versa. 

\begin{figure}[h]
\captionsetup[subfloat]{farskip=2pt,captionskip=1pt}
    \centering
    \subfloat[FGSM]{\includegraphics[width=2.8cm, height=2.8cm]{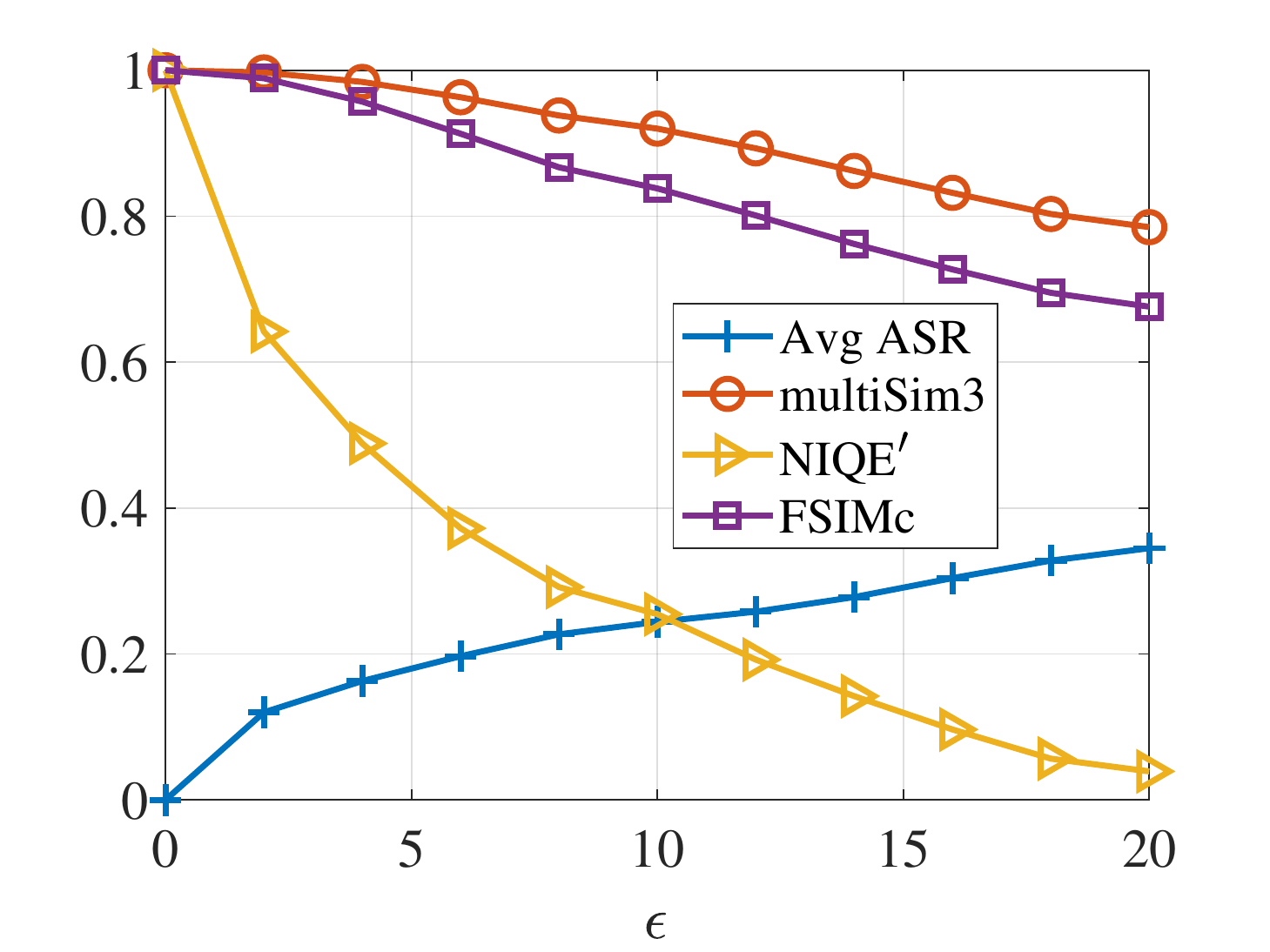} \label{fig:para_fgsm}}
    \subfloat[MIM]{\includegraphics[width=2.8cm, height=2.8cm]{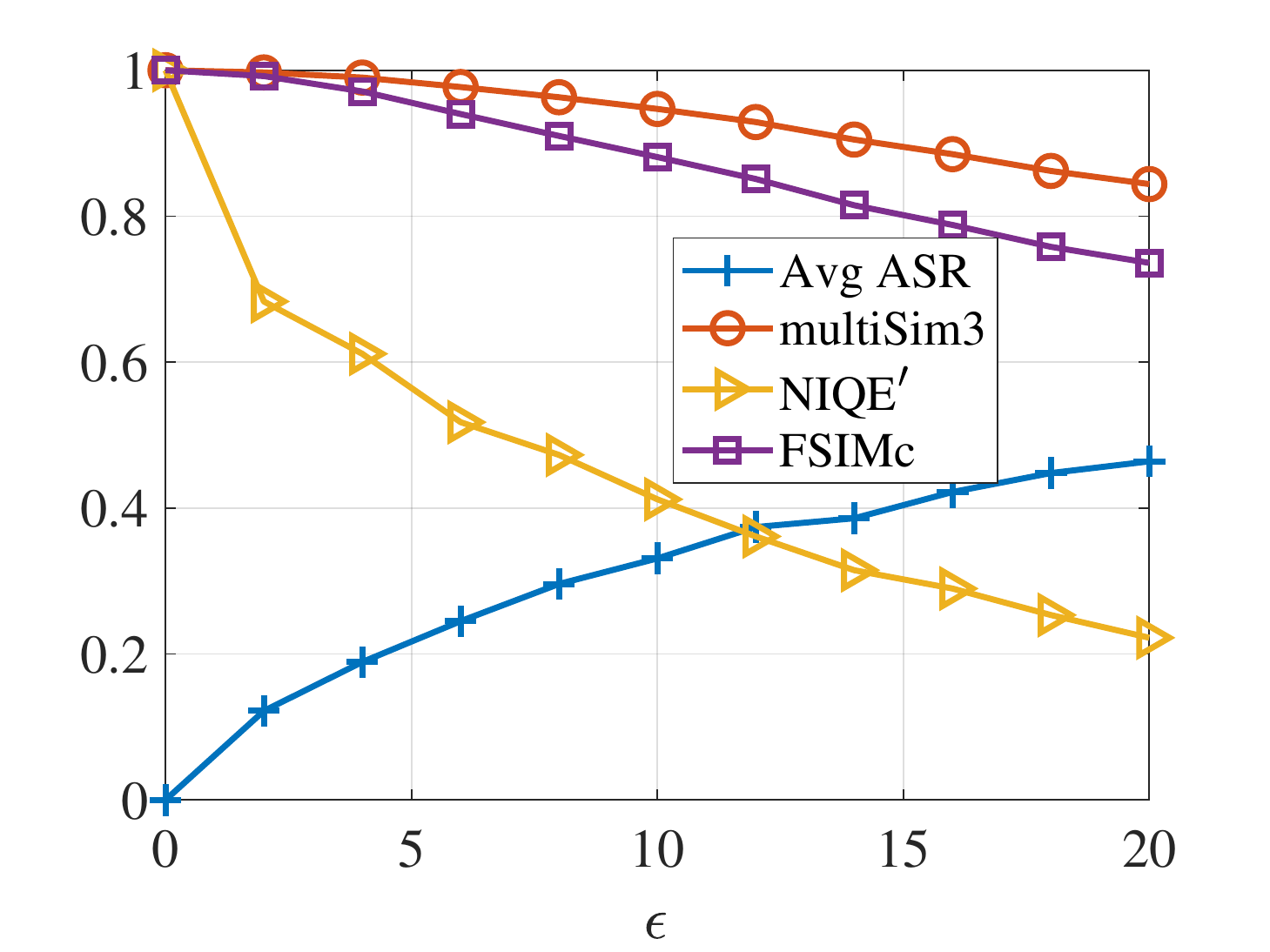} \label{fig:para_mim}}
    \subfloat[DIM]{\includegraphics[width=2.8cm, height=2.8cm]{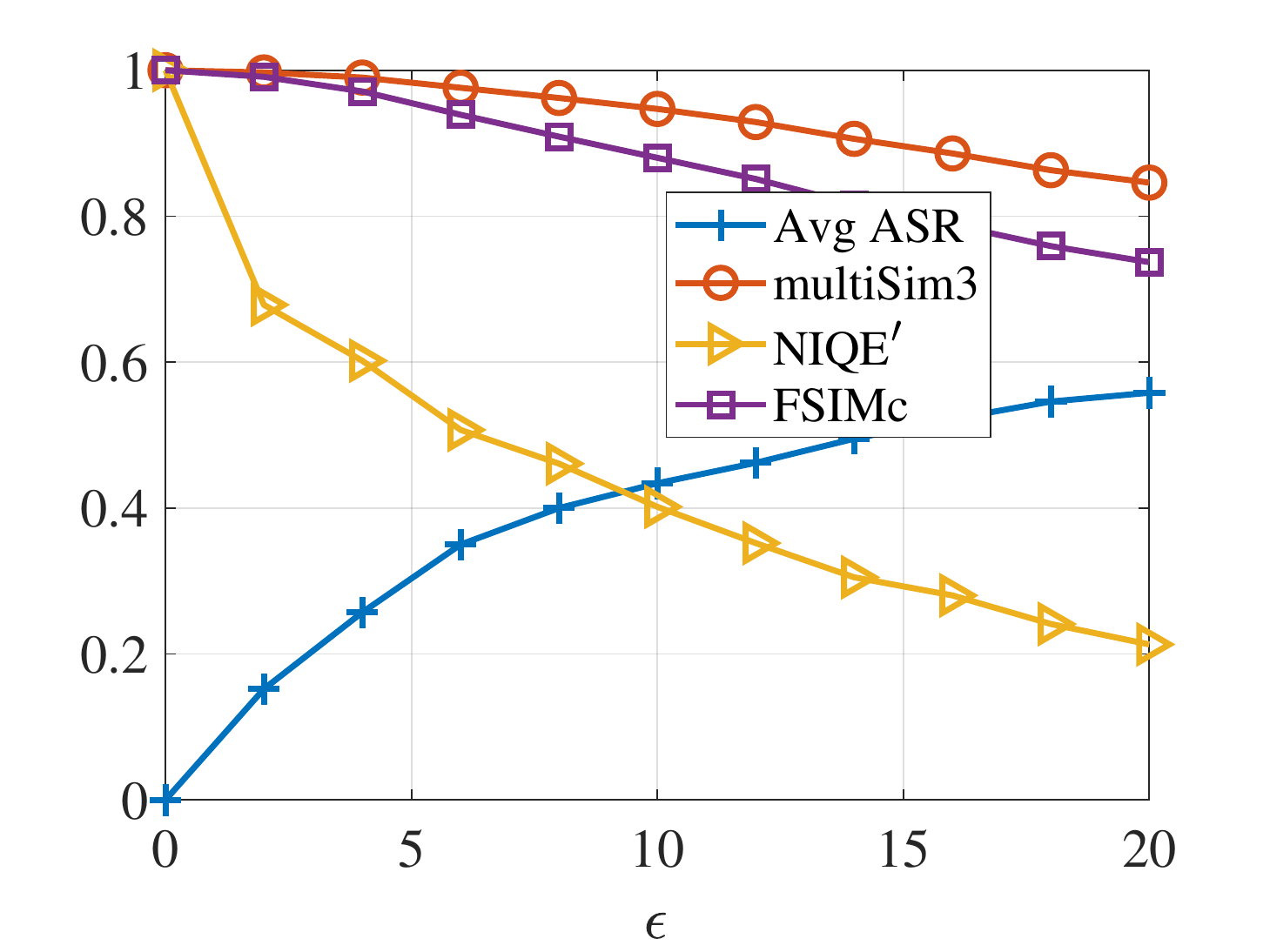} \label{fig:para_dim}} \\
    \subfloat[SSA-FGSM]{\includegraphics[width=2.8cm, height=2.8cm]{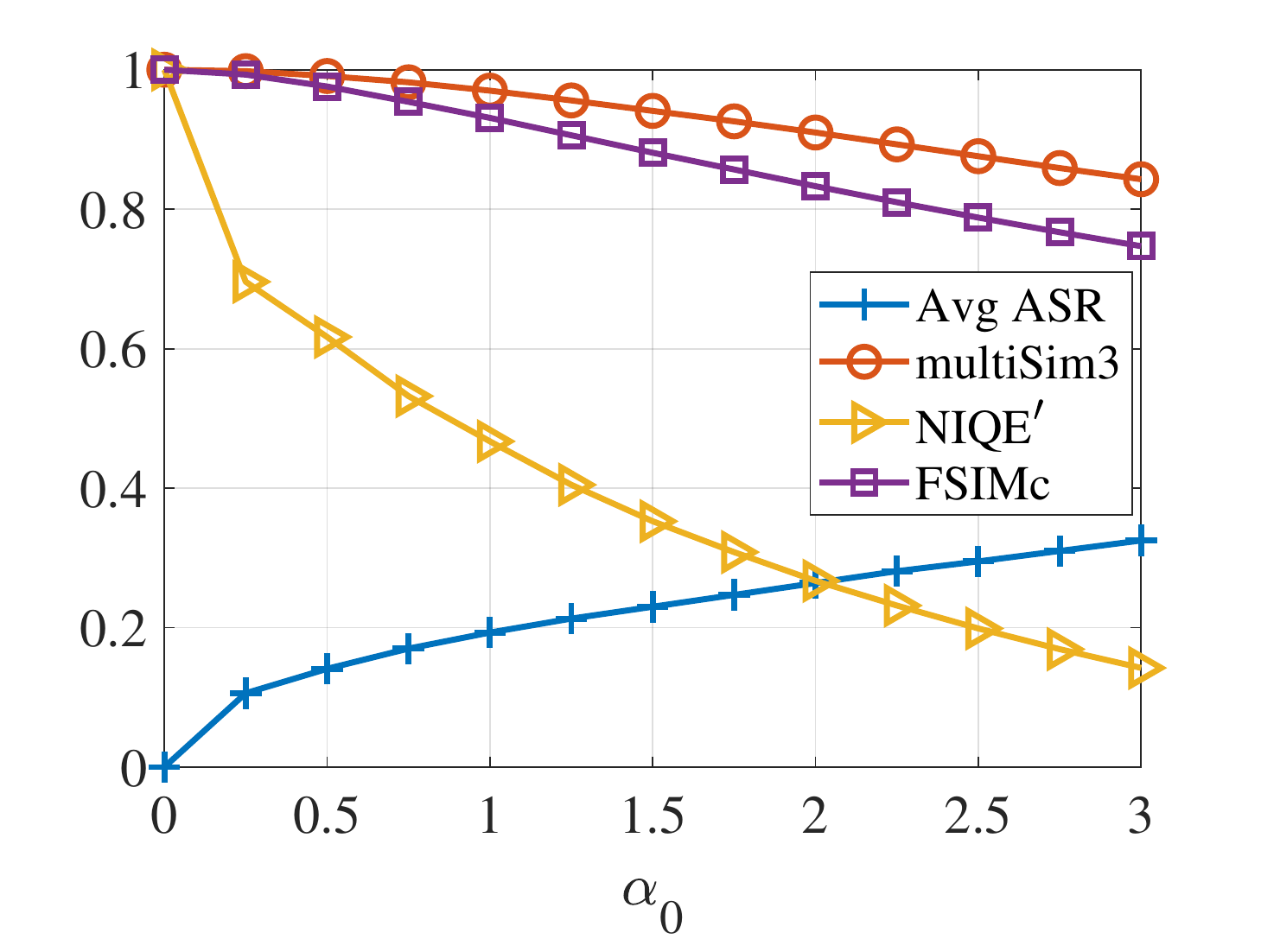} \label{fig:para_ssa_fgsm}} 
    \subfloat[SSA-MIM]{\includegraphics[width=2.8cm, height=2.8cm]{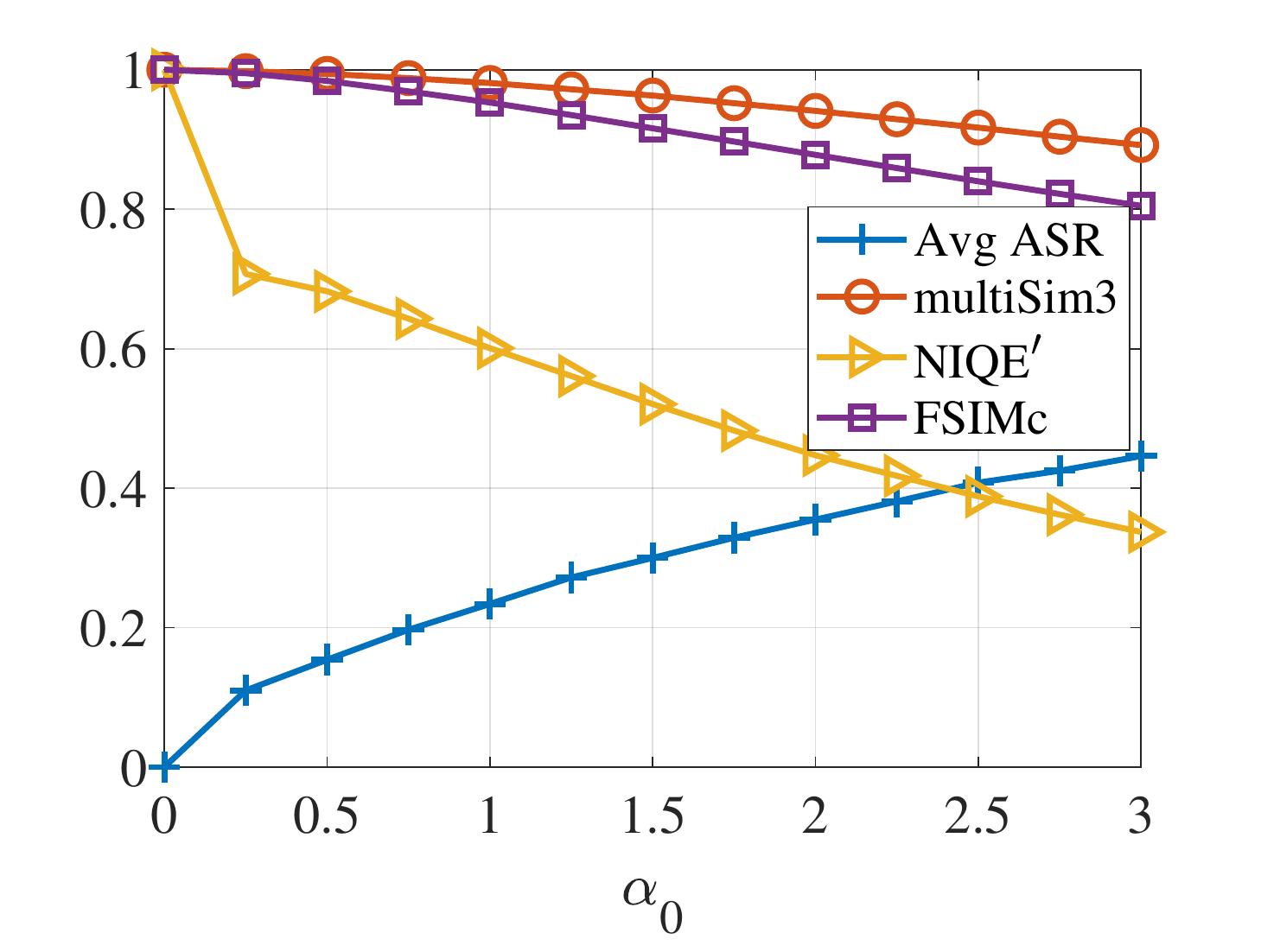} \label{fig:para_ssa_mim}} 
    \subfloat[SSA-DIM]{\includegraphics[width=2.8cm, height=2.8cm]{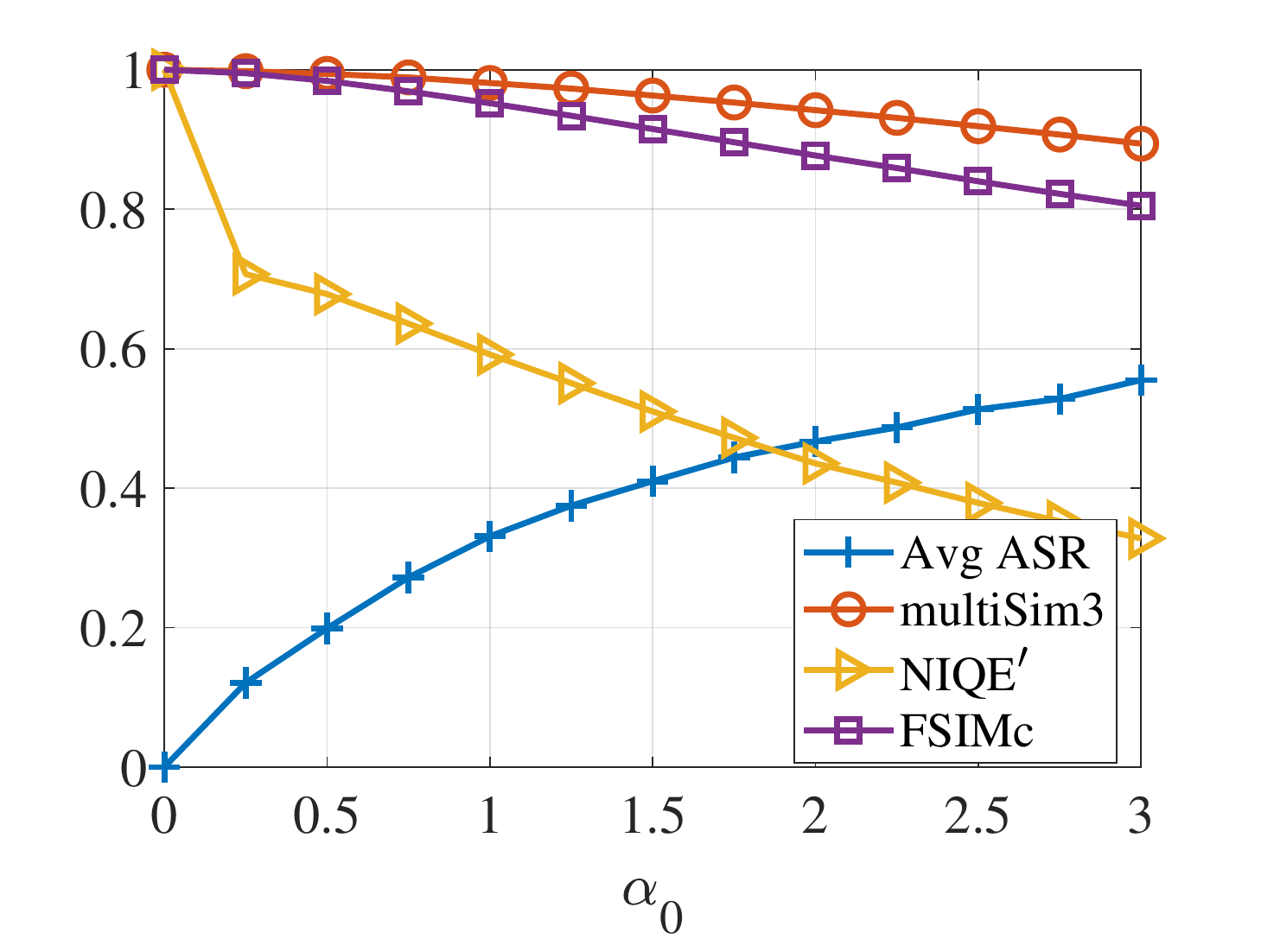} \label{fig:para_ssa_dim}} \\ 
    \subfloat[FSA-FGSM]{\includegraphics[width=2.8cm, height=2.8cm]{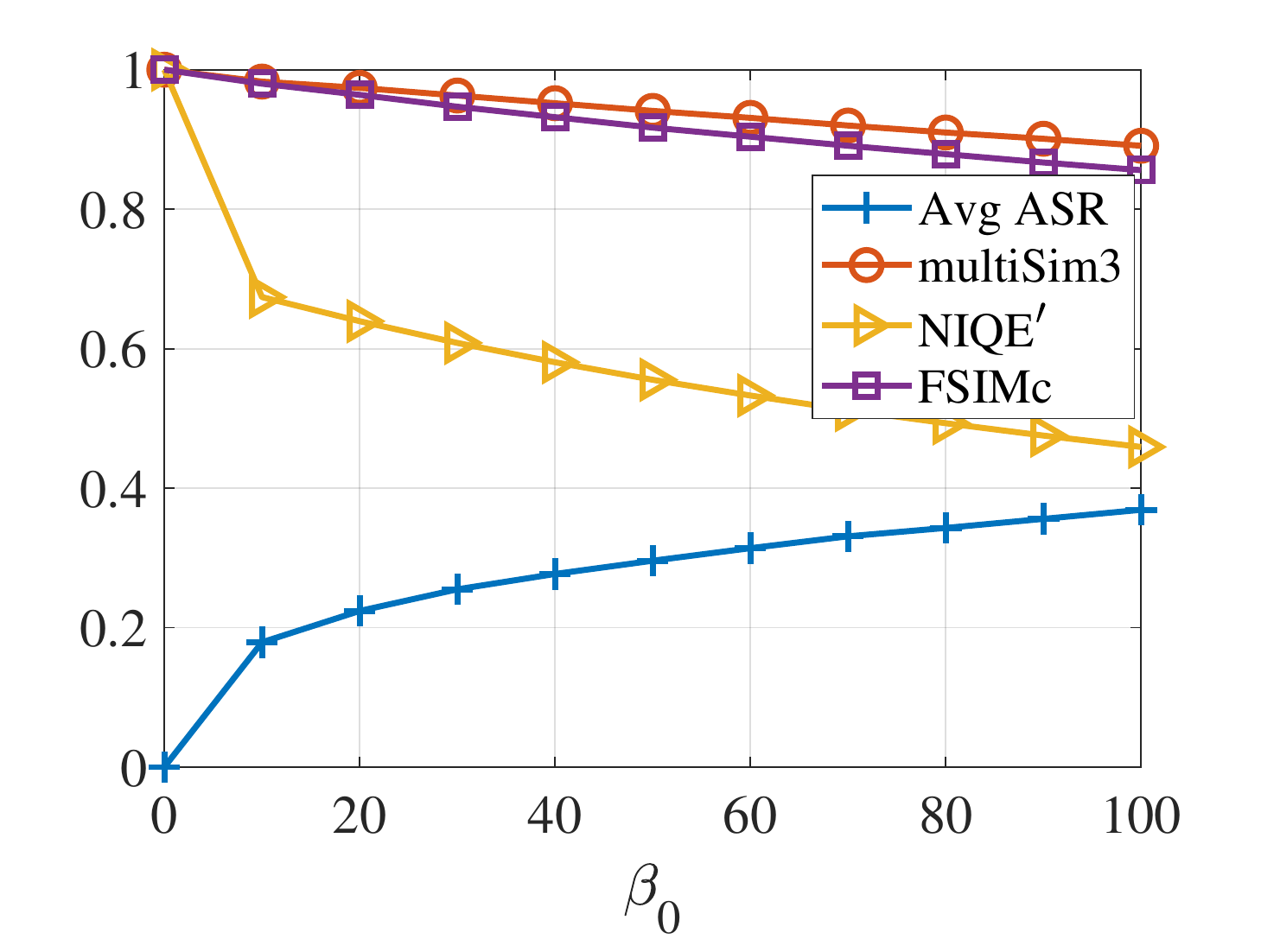} \label{fig:para_fsa_fgsm}} 
    \subfloat[FSA-MIM]{\includegraphics[width=2.8cm, height=2.8cm]{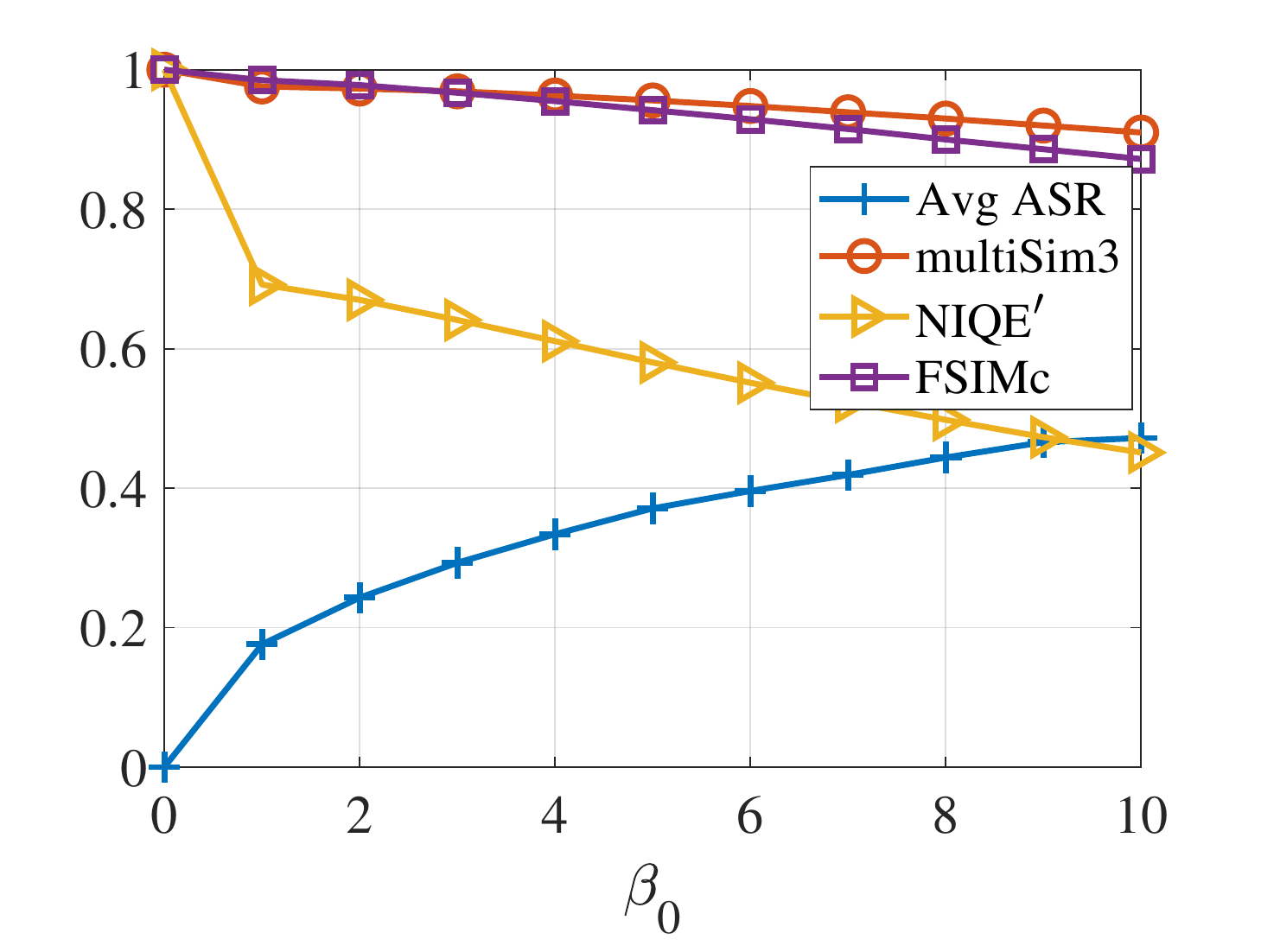} \label{fig:para_fsa_mim}}
    \subfloat[FSA-DIM]{\includegraphics[width=2.8cm, height=2.8cm]{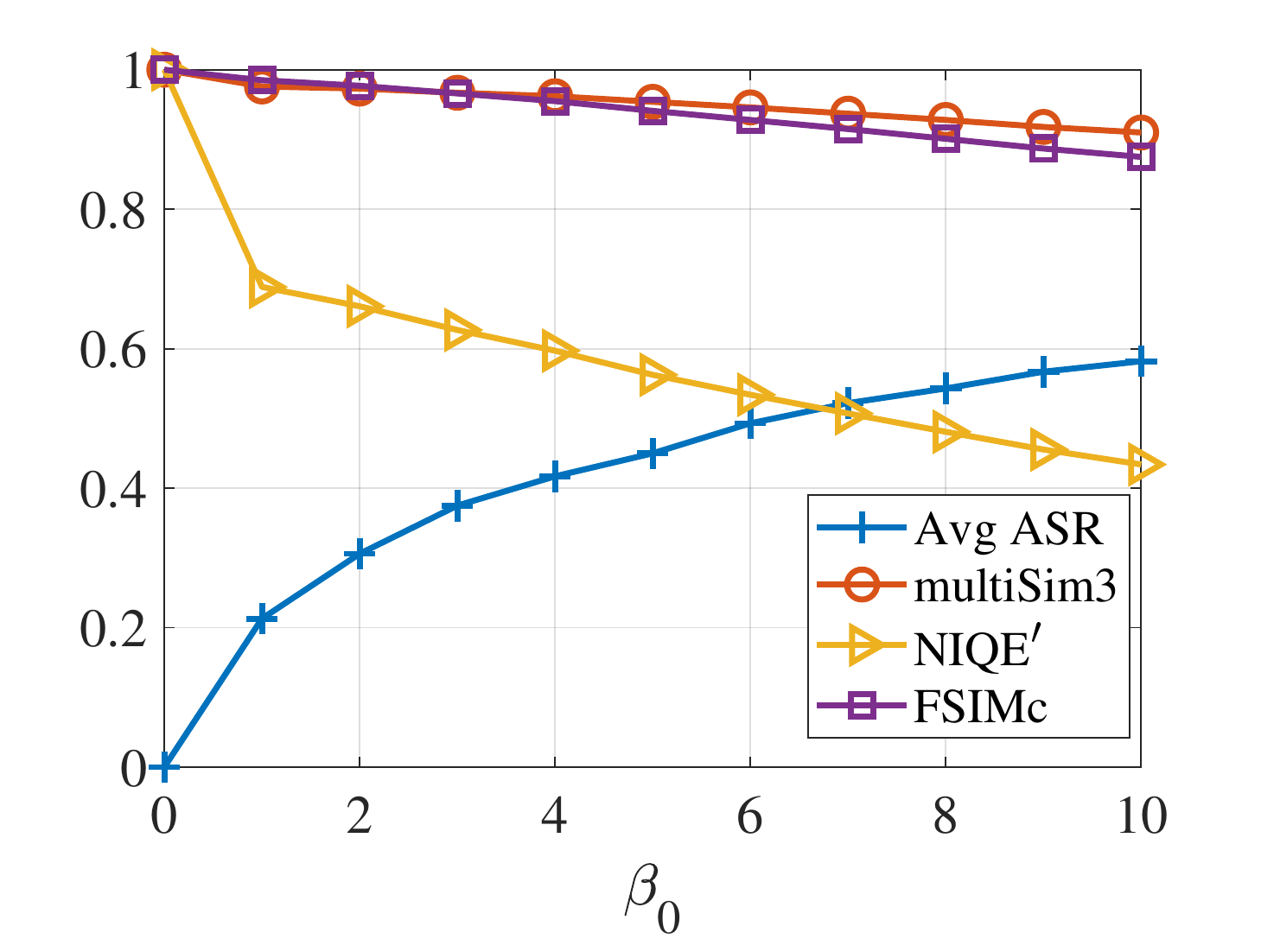} \label{fig:para_fsa_dim}}
    \caption{Parameter sensitivity comparisons between the baseline methods (i.e. subfigure (a)-(c)) and the proposed SSA (i.e. (d)-(f)) and FSA (i.e. (g)-(i)) based approaches.}
    \label{fig:para_sensitivity}
\end{figure}

In Fig.~\ref{fig:para_sensitivity}, we depict the parameter sensitivity curves for three baseline attacks (FGSM, MIM and DIM) and their perception-incorporated SSA/FSA based methods. The normalization factor $NIQE_{ub}$ equals 3.5 in the figures. In general, for each attack method, as hyperparameters (perturbation budgets) increase, the averaged ASRs increase at the expense of degraded visual quality (i.e. lower multiSim3, $\textrm{NIQE}'$ and FSIMc indices). We also observe that with comparable ASRs, the proposed methods consistently outperform their baselines. For instance, DIM achieves averaged ASR as 43.5\% at $\epsilon=10$ and its FSIMc equals 0.880. As a comparison, SSA-DIM produces averaged ASR to be 44.4\% with FSIMc as 0.896 at $\alpha_0=1.75$. Meanwhile, FSA-DIM attains its ASR as \textbf{45.0}\% with FSIMc equals \textbf{0.941} at $\beta_0=5.0$. The comparison results  answer the question that \textit{it is indeed possible to achieve a high ASR with improved visual quality}. 

\section{Conclusions and future work}
In this work, we present two novel approaches to improve the perceptual quality of adversarial examples for deep networks in the transfer-based black-box setting. Since the existing uniform perturbation constraint does not align well with human visual systems, we explicitly consider the regional and structural information of images and incorporate the perceptual models into adversarial attacks.  Specifically, we firstly introduce a spatial perceptual model and propose a structure-aware adversarial attack framework in the spatial domain. This framework is general and is compatible with all gradient-based attack methods. Further, we propose an adversarial attack framework by perturbing images in the frequency perceptual domain. Due to the structural constraints we explicitly consider, compared with baseline attacks, we demonstrate that adversarial examples produced by the proposed methods can generally have imperceptible or higher natural visual quality than the original attack methods with comparable attack success rates. Moreover, with comparable perceptual quality, the proposed methods produce higher attack success rates than baseline methods. In the future work, we plan to investigate and extend the proposed structure-aware frameworks to related tasks, e.g., imperceptible physical adversarial attacks.

\appendix
\section{Spatial JND Filters}
\label{app:spatial_JND}
In the spatial perceptual model (Section 3.1), the four high-pass oriented filters used in Eq.(3) are,
\begin{equation*}
    \centering 
    \scalemath{0.9}{
    \boldsymbol{h}_1= \frac{1}{16} \left[ \begin{matrix}
   0 & 0 & 0 & 0 & 0  \\
   1 & 3 & 8 & 3 & 1  \\
   0 & 0 & 0 & 0 & 0  \\
   -1 & -3 & -8 & -3 & -1  \\
   0 & 0 & 0 & 0 & 0  \\
\end{matrix} \right], 
\boldsymbol{h}_2= \frac{1}{16} \left[ \begin{matrix}
   0 & 0 & 1 & 0 & 0  \\
   0 & 8 & 3 & 0 & 0  \\
   1 & 3 & 0 & -3 & -1  \\
   0 & 0 & -3 & -8 & 0  \\
   0 & 0 & -1 & 0 & 0  \\
\end{matrix} \right]
}
\end{equation*}

\begin{equation*}
    \centering 
    \scalemath{0.95}{
    \boldsymbol{h}_3= \frac{1}{16} \left[ \begin{matrix}
   0 & 0 & 1 & 0 & 0  \\
   0 & 0 & 3 & 8 & 0  \\
   -1 & -3 & 0 & 3 & 1  \\
   0 & -8 & -3 & 0 & 0  \\
   0 & 0 & -1 & 0 & 0  \\
\end{matrix} \right], \;
\boldsymbol{h}_4= \frac{1}{16} \left[ \begin{matrix}
   0 & 1 & 0 & -1 & 0  \\
   0 & 3 & 0 & -3 & 0  \\
   0 & 8 & 0 & -8 & 0  \\
   0 & 3 & 0 & -3 & 0  \\
   0 & 1 & 0 & -1 & 0  \\
\end{matrix} \right]
}
\end{equation*}
The parameters for the Gaussian low-pass filer $\boldsymbol{l_g}$ used in Eq.(4) are: $3\times3$ Gaussian kernel with mean as 0 and standard deviation as 0.5.

\section{The MOS Test}
\label{mos_setting}
To test the perceptual improvement of our proposed framework, we design a subjective test for perceptual image quality evaluation.   We invite 10 volunteers to score the visual quality of the adversarial images. In each series of comparisons from Section \ref{sec:perceptual_improvement}, i.e. FGSM series, MIM series and DIM series, we randomly choose 50 adversarial examples from each adversarial attack method.  For example, in the FGSM series, we randomly select 50 adversarial images generated by FGSM, 50 images generated by SSA-FGSM, and 50 images generated by FSA-FGSM.  During the subjective test, we show a volunteer one pair of images and give her/him two seconds to review. The pair of images include an adversarial image and its corresponding clean image as reference. Finally the volunteer rates the adversarial image with a score. We repeat this process until all selected images are reviewed by this volunteer. The order of images for different volunteers are different. In this experiment, we employ the commonly used absolute category rating principle \cite{streijl2016mean}, with image quality score ranging from 1 to 5. The scores indicate:
\begin{itemize}
    \item score = 1: visually bad and very disturbing;
    \item score = 2: poor visual quality with disturbing visual artifacts;
    \item score = 3: fair visual quality with acceptable perceptual distortion;
    \item score = 4: good visual quality with slight perceptual distortion;
    \item score = 5: excellent visual quality with almost imperceptible distortion.
\end{itemize}
 
Mean opinion score (MOS) is computed by averaging subjective scores from all volunteers for each adversarial attack method.

\bibliographystyle{elsarticle-num} 
\bibliography{bib_file}


\end{document}